\documentclass{article}

\usepackage{arxiv}

\usepackage[utf8]{inputenc} 
\usepackage[T1]{fontenc}    
\usepackage{url}            
\usepackage{booktabs}       
\usepackage{amsfonts}       
\usepackage{nicefrac}       
\usepackage{microtype}      
\usepackage{algorithm}
\usepackage{algpseudocode}
\usepackage{multirow}
\usepackage{subfigure}

\usepackage{amsmath}
\usepackage{amssymb}
\usepackage{dsfont}
\usepackage{mathtools}
\usepackage{graphicx}
\usepackage{bm}
\usepackage{cancel}
\usepackage{url}
\usepackage{xcolor}
\newcommand{\citet}[1]{\cite{#1}}

\begin{document}

\title{Jacobian Granger Causal Neural Networks for Analysis of Stationary and Nonstationary Data}

\author{Suryadi\textsuperscript{1}, Yew-Soon Ong\textsuperscript{2}, Lock Yue Chew\textsuperscript{1} \\ 1: School of Physical and Mathematical Sciences, Nanyang Technological University, Singapore. \\2: School of Computer Science and Engineering, Nanyang Technological University, Singapore}
\date{}
\maketitle

\begin{abstract}
Granger causality is a commonly used method for uncovering information flow and dependencies in a time series. Here we introduce JGC (Jacobian Granger Causality), a neural network-based approach to Granger causality using the Jacobian as a measure of variable importance, and propose a thresholding procedure for inferring Granger causal variables using this measure. The resulting approach performs consistently well compared to other approaches in identifying Granger causal variables, the associated time lags, as well as interaction signs. Lastly, through the inclusion of a time variable, we show that this approach is able to learn the temporal dependencies for nonstationary systems whose Granger causal structures change in time.
\end{abstract}

\section{Introduction} The evolution of an observed system usually occurs with some definite and possibly nonstationary couplings between its constituents. As a starting point, information flow allows one to begin uncovering such a structure, which also serves as a precursor to a causal understanding of the system. Among other methods, Granger causality (GC)  \cite{originalGC,GC_inf_flow, GC_info_theory} has been used for such a purpose in various fields including finance and economics \cite{finance1,econ1,finance2,finance3}, neuroscience \cite{brain_info_flow,GC_neuroscience,GC_fMRI,Reid}, climate and environmental studies \cite{env1,env2,env3}, as well as biological networks \cite{bio1,bio2,bio3}. More generally, GC can be framed as feature selection in multivariate time series \cite{feature_selection}. Another common approach to information flow is the transfer entropy and its extensions \cite{original_transfer_ent,te_spo}, which has also been shown to be equivalent to Granger causality under some conditions \cite{GC_transfer_entropy, GC_transfer_generalized}.

Recent efforts on Granger causality worked towards extending it to high dimensional and nonlinear data in a model-free manner using neural networks without the need for exponentially large data size. These developments were extensively tested for their ability to infer the correct Granger causal variables. While some of these approaches are also capable of extracting the time lag for each discovered interaction, no prior study was conducted to specifically test the ability to correctly infer the interaction time lag. This information can be useful for theoretical modeling and understanding of the system at hand, and it is also studied in the field of dynamical systems \cite{delay}. In addition, it may also provide information for confounder analysis \cite{bahadori2013examination}.

In addition, these approaches (including information flow measures such as the transfer entropy) require the structure of the data to be stationary, i.e. the interaction between variables in the system do not change in time. We see this being violated in systems with structural breaks in financial modeling \cite{structural_breaks}, as well as in neuroscience where the system may shift between different functional networks in a single time series \cite{safikhani2020joint}.

The contributions of this work are as follows:
\begin{enumerate}
    \item We augment the model for Granger causality to include contemporaneous variables. While they have not been widely used in most of the recent related works on Granger causality using neural networks, we provide an argument to illustrate its importance.
    \item We propose the use of the Jacobian with respect to input as a measure of variable importance to infer Granger causality. We denote this approach as Jacobian Granger Causality (JGC).
    \item We compare the proposed approach with other neural network-based methods for Granger causality across multiple test cases. In particular, we consider separately the inference of Granger causal variables, Granger causal variables with time lag, as well as effect sign.
    \item Lastly, we consider the problem of nonstationary data whose structure changes with time. By augmenting the model with a measure of time,  we show empirically that the proposed approach is capable of discovering these structural changes.
\end{enumerate}

\section{Background and Related Work}
\subsection{Granger Causality}
Given an N-dimensional time series $\{x_i(t)\}_{i=1,\ldots,N}$ with time represented in units of integer timesteps and a specific target variable $x_j(t)$, we want to find the smallest possible subset of $\{x_i(t-\alpha)\}_{i=1,\ldots,N;\alpha=1,\ldots,\eta}$ such that
\begin{equation}\label{GC_eq}
    x_j(t)=f_j(x_1(t-1),\ldots,x_k(t-\tau))+\epsilon_j
\end{equation}
where $\tau\leq\eta$ with $\tau$ representing the maximum interaction time lag discovered in the system, and $\eta$ the maximum lag tested by the procedure. $\epsilon_j$ is a noise term. The specification of $\eta$ may be informed by prior knowledge of the system and may also be constrained by the amount of data available. In any case, $\eta$ should be sufficiently large in order to capture all the lagged interactions. We designate the minimal set of variables $\{x_i\}_{i=1,\ldots,k}$ in the right hand side of \eqref{GC_eq} as the Granger causal variables for $x_j$ and the remaining variables as irrelevant variables. In other words, all variables in the set must contribute to the prediction of the target variable given the information about all other variables in the set. 

Granger causality can be represented as a directed graph \cite{eichler} with the vertices representing the variables and edges representing the directed relationship from the Granger causal variables to the corresponding target variables. On the other hand, the interaction time may also be incorporated into the graph by labeling the edges \cite{bahadori2013examination}. While it renders the graph more informative, it also allows the determination of constraints that confounders must satisfy, if they exist. 

\subsection{Related Work}
Recent years saw the development of multiple approaches of using neural network to infer Granger causality. Initially conceived in 2018, \citet{tank} saw the use of MLP and LSTM, denoted cMLP and cLSTM, where inference is done by a novel loss function with sparsity constraints and optimized by proximal gradient descent. Granger causality is then made unambiguous as all irrelevant variables have zero weights in the first hidden layer. This approach was subsequently extended by \citet{esru} through the use of statistical recurrent units (SRU), dubbed economy-SRU (eSRU) due to its sample-efficient property. Convolutional nets were also used, in an approach called temporal causal discovery framework (TCDF), where attention scores were used to determine the relevance of variables \cite{tcdf}. More recently, \citet{gvar} extended self-explaining neural networks towards Granger causality by modeling the system as a generalized vector autoregression (GVAR), in which case the coefficients in GVAR measures the relevance of the variables. Unlike the other mentioned approaches, TCDF and GVAR return nonzero measures of variable importance, hence a subsequent thresholding procedure was required to infer Granger causal variables. TCDF utilizes a permutation test, while GVAR compares with the analysis when the data is given in reverse temporal order. The latter works on the premise of stability \cite{stability}, where the true Granger causal variables are expected to be consistent across different runs. Out of these approaches, information on the interaction time lag could be extracted from cMLP, TCDF, and GVAR.

\section{Method}
\subsection{Augmenting the Granger Causal Model}\label{augment}
We begin with a fresh formulation of Granger causality. As discussed in \eqref{GC_eq}, formulations of Granger causality typically consider candidate variables at some nonzero delay with respect to the target variable. Here, we augment the model by adding contemporaneous variables, which are the other variables measured at the same time as the target variable that we are interested to predict. If the target variable is $x_j(t)$, then the contemporaneous variables are $\{x_i(t)\}_{i\neq j}$.  With this, \eqref{GC_eq} is rewritten as
\begin{equation}\label{augGC_eq}
    x_j(t)=f_j(x_i(t),\ldots,x_k(t-\tau))+\epsilon_j
\end{equation}
with $i\neq j$. To motivate this augmentation, we consider a simple system evolving under a nonlinear differential equation with some function $g$,
\begin{equation}\label{z_eq}
    \frac{dz(t)}{dt}=g(x(t))
\end{equation}
Since we are dealing with data measured at discrete timesteps, we can rewrite \eqref{z_eq} in terms of the given data as 
\begin{equation}\label{z_update}
    z(t)=z(t-1)+\int_{(t-1)}^{t}g(x(t'))\ dt'
\end{equation}
Denoting the antiderivative of $g$ as $G$, we then have
\begin{equation}\label{z_final}
    z(t)=z(t-1)+G(x(t)) - G(x(t-1))
\end{equation}
We emphasize from \eqref{z_eq} that since we are testing for Granger causality, the function $g$ should contain only the direct causes of $z$. However, as \eqref{z_final} indicates, information on both $x(t)$ and $x(t-1)$ are required to accurately predict $z(t)$. In the absence of $x(t)$, i.e. given only $x(t-1)$ and $z(t-1)$, accurate prediction of $z(t)$ would then necessitate modeling the evolution of $x(t)$ as well, in which case $g$ potentially requires causes of $x$ that are not causes of $z$, which may lead to false positives. Neglecting contemporaneous variables (where such causality exists) also has an impact on autoregressive systems, which has been discussed at length in \cite{contemporaneous}. Lastly, we note that contemporaneous variables were in fact incorporated in TCDF, although other more recent approaches do not make use of them. 

\subsection{Jacobian as a Measure of Variable Importance}\label{sec3.2}
\begin{figure}[b]
    \centering
    \includegraphics[width=1.6in]{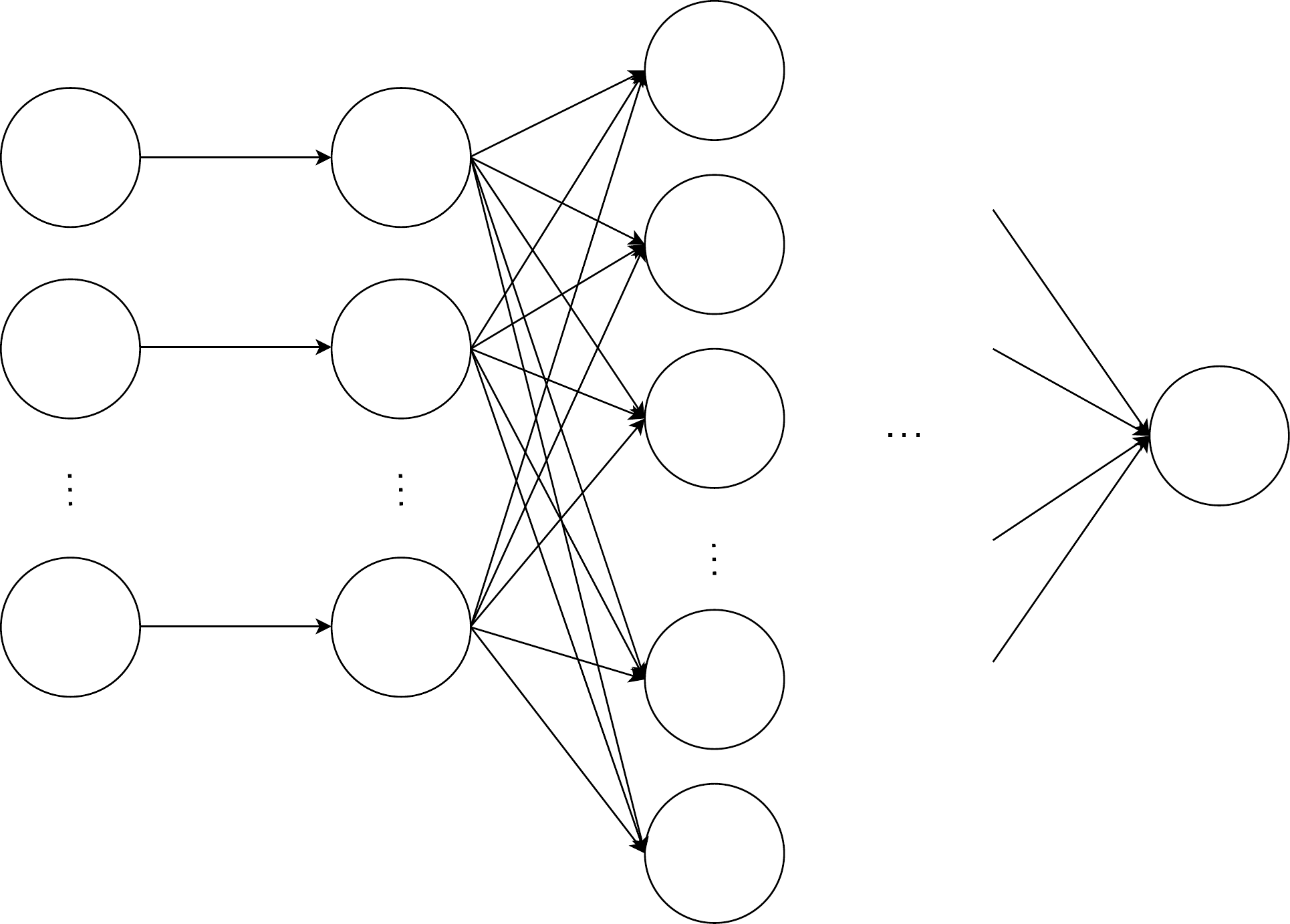}
    \caption{The neural network architecture for nonlinear Granger causality.}
    \label{fig:1}
\end{figure}

We approximate the unknown function $f_j$ in \eqref{augGC_eq} using a neural network. The architecture of the network is a simple feedforward architecture with the exception of the first hidden layer. This layer is not fully connected from the input layer, but is instead wired one-to-one (Fig. \ref{fig:1}). This allows the first hidden layer to serve as a gate for input information through its weights. We can therefore bias the network into using only the information from relevant inputs. Given $N$ variables and a predetermined maximum test lag $\eta$, each variable at each lag is fed into the network as a separate input, hence the dimension of the input layer is $D=N\eta$. We note that a separate neural network would be required for each target variable.

The network is then trained in a supervised manner with the target variable $x_j(t)$ as the output. The training is then done with the loss function defined as $MSE+\sum_k\lambda|w_k|$ where $w_k$ is the weight of the $k$th neuron in the first hidden layer. 

From \eqref{augGC_eq}, the contribution of some variable $x_i(t-\alpha)$ at a given lag $\alpha$ on the evolution of $x_j(t)$ can be written to first order as the Jacobian with respect to the input,
\begin{equation}\label{diff}
    \frac{\partial x_j(t)}{\partial x_i(t-\alpha)} = \frac{\partial f_j}{\partial x_i(t-\alpha)} =: f'_{j,i\alpha}
\end{equation}
With the trained neural network serving as a representation of $f_j$, we can then use the Jacobian of the trained network with respect to each input as a measure of variable importance. This quantity then serves as an estimator of the true Jacobian in \eqref{diff}, in the same way that the neural network serves as an approximation of the unknown function $f_j$. This is a more holistic measure compared to weights as it summarizes the contribution of each variable over all layers. We denote the estimator of \eqref{diff}, as computed from the neural network, as $\hat{f}'_{j,i\alpha}$. Since the dataset consists of time series data, each input $x_i(t-\alpha)$ is indexed by time $t$, which therefore yields a vector of derivatives $\hat{f}'_{j,i\alpha,t}$ also indexed by $t$, which serves as a measure of variable importance. This then naturally allows us to visualize how the importance of each variables evolves in time. However, since information on time is not given to the network, it is unable to learn the time dependence of the variable importance measures, thereby encoding an implicit assumption where the importance is a function of all the other variables $x_i$ but not $t$ explicitly. This is true only if the dependency structure is stationary. We will address the nonstationary case in section \ref{nonstat}.

\subsection{Stationary Case: Inferring the Granger Causal Variables}\label{sec3.3}
Under the assumption of a stationary structure, which is typically the case in Granger causality, we can infer the Granger causal structure by first condensing the importance measures $\hat{f}'_{j,i\alpha,t}$ into a single summary statistic. We achieve this by computing the absolute mean value $|\overline{\hat{f}'_{j,i\alpha}}|$, which summarizes the contribution of each input. As the discussion in section \ref{augment} indicates, dependencies that occur without a time lag (eq. \eqref{z_eq}) may nevertheless require information on the relevant variables at a time lag of 1 timestep (eq. \eqref{z_final}), owing to the process of measurement which occurs at discrete time. Therefore, the contribution of a variable at a time lag of 1 timestep may in fact indicate dependencies at zero lag. This is also how other approaches with no contemporaneous variables measure such dependencies (e.g. in the Lorenz96 system in sections \ref{Lorenz} and \ref{sec4.2.1}). Since both quantities contribute towards the same conclusion, we combine $|\overline{\hat{f}'_{j,i\alpha}}|$ at lags of zero and one into a single number by summing them. We therefore do not distinguish between dependencies with lags of zero and one.

With this, we can then identify the Granger causal variable by applying an appropriate threshold. While a good threshold can in fact sometimes be identified by eye, we discuss here a procedure that allows a more systematic way for determining the threshold. The procedure operates under the assumption that the value of $|\overline{\hat{f}'_{j,i\alpha}}|$ is consistently larger for the Granger causal variables relative to the irrelevant ones. While the Granger causal variables consistently contribute to the prediction, the contributions from the irrelevant variables should essentially be noisy and therefore inconsistent in their ordering. This idea is similar to the basis for stability selection \cite{stability}.

To make use of these properties, we run multiple independent runs of the procedure, which can be run in parallel; in our implementation we make use of 3 independent runs.

To simplify the subsequent step, variables with negligible contributions are first removed, which also reduces the false positive rates. There can be multiple ways for performing such a step; here we simply remove all variables whose $\overline{|\hat{f}'_{j,i\alpha}|}$ is less than $0.01$ of the largest value in the respective run; a different cut-off may be chosen if required. In a high dimensional system with sparse couplings, this step should remove most of the tested variables, leaving a smaller set of potential Granger causal variables for each run.

Subsequently, we compute the intersection among all the sets from respective runs. Each set is then modified by imposing an additional condition where we only retain a variable $x_k(t-\beta)$ in each set if all other variables in the set with $\overline{|\hat{f}'_{j,i\alpha}|}$ larger or equal than that of $x_k(t-\beta)$ in each set are in the intersection. The intersection of the sets is taken again and this procedure is repeated once. The final intersection of these modified sets then constitute the inferred Granger causal variables at the respective time lag. This procedure is summarized in algorithm \ref{alg:thres}. For ease of subsequent comparisons, we refer to our proposed method outlined here in sections \ref{sec3.2} and \ref{sec3.3} as Jacobian GC (JGC).

\begin{algorithm}[tb]
   \caption{Thresholding procedure for inferring Granger causal variables}
   \label{alg:thres}
\begin{algorithmic}
   \State {\bfseries Input:} $s_1,s_2,s_3$ (sets of variable importance scores for all input for target variable $x_j$ from 3 runs), cutoff $\epsilon$ 
   \State {\bfseries Output:} GC, the inferred Granger causal variables for $x_j$ with time lag information
   \For{$i=1$ {\bfseries to} $3$}
   \State set to zero all elements of $s_i<\epsilon\cdot max(s_i)$ 
   \State $o_i=argsort(s_i)$ in descending order
   \EndFor
   \For{$n=1$ {\bfseries to} $2$}
   \State $I=intersection(o_1,o_2,o_3)$
   \For{$i=1$ {\bfseries to} $3$}
   \For{$k=1$ {\bfseries to} length($o_i$)}
   \If{$o_{i,k}\notin I$}
   \State $o_{i} = [o_{i,1},o_{i,2},\ldots,o_{i,k-1}]$
   \State break
   \EndIf
   \EndFor
   \EndFor
   \EndFor
\State $GC=intersection(o_1,o_2,o_3)$
\State {\bfseries return} GC
\end{algorithmic}
\end{algorithm}

\subsection{Nonstationary Case: Visualizing Instantaneous Information Flow}\label{nonstat}
The formulation in \eqref{augGC_eq} expresses $x_j(t)$ as a function of all other variables $x_i$ and not $t$ explicitly, encoding the assumption that $f_j$ (and therefore the dependency structure) does not change directly with time. To handle systems where this assumption does not hold, we propose to augment $f_j$ with a measure of time such that there is a definite way of distinguishing the temporal order of the data, thereby allowing the model to adapt its connections and their strengths as a function of time. The model then becomes
\begin{equation}\label{nonstat_eq}
    x_j(t)=f_j(x_i(t),\ldots,x_k(t-\tau),t)+\epsilon_j
\end{equation}
Under this formulation, the network can then learn how the importance measures $\hat{f}'_{j,i\alpha,t}$ change with time, and this time dependence can then be visualized.

Experiments indicate that the inclusion of time may lead to slightly worse performance for Granger causality when the causal structure has no time dependence. Therefore, we suggest that a measure of time can be added when time-dependence is suspected, after which the time-dependence could be determined (as seen in section \ref{nonstationary}). This then serves as a stationarity test. If no significant time-dependence in the dependencies is observed, i.e. if the variable importance measures do not appear to vary significantly with time, the procedure may still proceed without the time variable.

\section{Experiments on Stationary Granger Causality} 
In this section, we examine how the introduced approach performs in different simulated systems where the true Granger causal structure is known. In addition, the dependency structure is assumed to be stationary as per the standard Granger causality setting. We then compare it with the other known approaches through two scores, namely the area under receiver operating characteristic curve (AUROC) and the area under precision-recall curve (AUPRC). Each method was run through a range of sparsity hyperparameter values and the hyperparameters corresponding to the highest average AUPRC were taken for comparison, similar to \citet{gvar}. The reported score for each test case was averaged over 5 independently generated realizations of the system. The hyperparameters associated with each method for each test case are given in the appendix. Lastly, the results for JGC over all experiments in this section are computed without augmenting the model with time. 

We organize this section into five subsections, the first three comparing the different approaches on the identification of Granger causal variables, time lag, and interaction sign respectively. The fourth subsection averages the results across the first three subsection for an overall comparison of the top performing models.
As AUROC and AUPRC are computed without thresholding the variables, we evaluate the thresholding procedures by instead analyzing their stability relative to the regularization parameter(s) in the fifth subsection. Computations were run on a computer with AMD Ryzen 7 3700X 8-core processor (3.59 GHz) and NVIDIA GeForce RTX 2070 SUPER. The code for the implementation and data used for these experiments are available at [link].

\subsection{Identification of Granger Causal Variables}\label{sec4.1}
In this section, we consider the identification of the Granger causal variables, ignoring time lag information. In other words, the identification of the correct variable at the wrong time lag is considered correct. We describe each tested system below, while the results are collated in Table \ref{table1}.

\subsubsection{Vector Autoregression}\label{varsection}
We begin by considering vector autoregressions (VARs) as defined below with dimension $N=10$:
\begin{equation}\label{VAR}
    \mathbf{x}(t) = \sum_{\alpha=1}^\tau A_\alpha \mathbf{x}(t-\alpha) + \bm\epsilon
\end{equation}
Here $A_\alpha$ denotes the connectivity matrix of the system acting at a delay of $\alpha$ and $\bm\epsilon$ is a noise vector with each element $\sim N(0,1)$. The system is set to be sparsely connected with random connections, hence any pair of variables can potentially be connected at only a single or multiple arbitrary time lags. For this system, we simulated up to $T=500$ observations with the maximum lag $\tau=5$.

Table \ref{table1} indicates that cMLP, GVAR, and JGC performed very well for this dataset, while the remaining approaches are weaker. This is likely because the causal connections in this test case were randomly generated across multiple lags, hence there are many connections that only exist at larger lags, which may not be as easily detected by some of the tested approaches.

\subsubsection{Lorenz96}\label{Lorenz}
The next test case is the Lorenz96 system \cite{tank} with the evolution defined as:
\begin{equation*}
    \frac{dx_i(t)}{dt} = \left[x_{i+1}(t)-x_{i-2}(t)\right]x_{i-1}(t)-x_i(t)+ F,
\end{equation*}
with $i=1,\ldots,N$ where $F$ is a driving constant and the interaction defined symmetrically with $x_{-1}=x_{N}$. Here we simulated the system with dimension $N=20$ and sampled at a temporal sampling rate of $\Delta_t=0.1$. Similar to the baselines used for the other approaches, we consider two cases at $F=10,40$. As above, we simulated this system up to $T=500$. As is the case in \citet{gvar}, we exclude the trivial self-connections when computing the scores for this system.

We see that cMLP, eSRU, GVAR, and JGC all performed well for the two cases of Lorenz96, while cLSTM did significantly worse for the $F=40$ case compared to $F=10$. While cMLP has the highest scores for the $F=40$ case, we see later in section \ref{sec4.5} that the F-score of the thresholded results perform worse compared to GVAR and JGC.

\subsubsection{fMRI Simulations}
Lastly, we consider the realistic and highly nonlinear fMRI simulations with blood-oxygen-level-dependent (BOLD) signals \cite{fmri}. As the name suggests, the signals in the timeseries involves modeling both the brain activity as well as the associated vascular dynamics. Similar to the baselines in the other approaches, we use the first 5 samples from the third simulation of the original dataset, with dimension $N=15$ and length $T=200$ for each sample. As with Lorenz96 above, we exclude self-connections for this system.

For this test case, JGC and TCDF have significantly better performance compared to the rest. As these two approaches are the only ones with contemporaneous variables, the improvement is likely due to this factor.

\begin{table}[t] 
\caption{Comparison of our proposed approach (JGC) with other existing approaches in identifying Granger causal variables over multiple datasets: VAR, Lorenz96 ($F=10,40$), and fMRI.} 
\label{table1} 
\vskip 0.15in 
\centering
\begin{small} 
\begin{sc} 
\begin{tabular}{llcc} 
\toprule 
Dataset & Model & AUROC ($\pm$ SD) & AUPRC ($\pm$ SD) \\ 
\midrule 
\multirow{6}*{VAR}& cMLP & \textbf{0.999 $\pm$ 0.001} & \textbf{0.998 $\pm$ 0.002} \\ 
& cLSTM & 0.719 $\pm$ 0.099 & 0.652 $\pm$ 0.109 \\ 
& TCDF & 0.882 $\pm$ 0.027 & 0.851 $\pm$ 0.025 \\ 
& eSRU & 0.551 $\pm$ 0.031 & 0.519 $\pm$ 0.015 \\ 
& GVAR & 0.996 $\pm$ 0.003 & 0.994 $\pm$ 0.005 \\  
& JGC & 0.998 $\pm$ 0.003 & 0.997 $\pm$ 0.003 \\
\midrule 
\multirow{6}*{L96$_{F=10}$}& cMLP & 0.965 $\pm$ 0.014 & 0.938 $\pm$ 0.024 \\ 
& cLSTM & 0.986 $\pm$ 0.005 & 0.969 $\pm$ 0.003 \\ 
& TCDF & 0.862 $\pm$ 0.017 & 0.645 $\pm$ 0.029 \\ 
& eSRU & 0.999 $\pm$ 0.001 & 0.995 $\pm$ 0.003 \\ 
& GVAR & \textbf{1.000 $\pm$ 0.000} & \textbf{0.999 $\pm$ 0.001} \\ 
& JGC & \textbf{1.000 $\pm$ 0.000} & \textbf{0.999 $\pm$ 0.001} \\ 
\midrule 
\multirow{6}*{L96$_{F=40}$}& cMLP & \textbf{0.996 $\pm$ 0.006} & \textbf{0.991 $\pm$ 0.011} \\ 
& cLSTM & 0.656 $\pm$ 0.042 & 0.390 $\pm$ 0.065 \\ 
& TCDF & 0.645 $\pm$ 0.031 & 0.312 $\pm$ 0.035 \\ 
& eSRU & 0.963 $\pm$ 0.013 & 0.878 $\pm$ 0.031 \\ 
& GVAR & 0.978 $\pm$ 0.004 & 0.935 $\pm$ 0.016 \\ 
& JGC & 0.970 $\pm$ 0.021 & 0.929 $\pm$ 0.027 \\ 
\midrule 
\multirow{6}*{fMRI}& cMLP & 0.605 $\pm$ 0.073 & 0.187 $\pm$ 0.057 \\ 
& cLSTM & 0.585 $\pm$ 0.060 & 0.169 $\pm$ 0.047 \\ 
& TCDF & 0.865 $\pm$ 0.018 & 0.431 $\pm$ 0.054 \\ 
& eSRU & 0.636 $\pm$ 0.103 & 0.186 $\pm$ 0.057 \\ 
& GVAR & 0.668 $\pm$ 0.092 & 0.257 $\pm$ 0.069 \\ 
& JGC & \textbf{0.886 $\pm$ 0.024} & \textbf{0.438 $\pm$ 0.086} \\ 
\bottomrule 
\end{tabular} 
\end{sc} 
\end{small} 
\vskip -0.1in 
\end{table} 

\subsection{Identification of the Interaction Time Lag}\label{sec4.2}
Here we test the correct identification of the interaction time lag in addition to the Granger causal variable itself. In other words, a Granger causal variable identified at the wrong time lag is considered a false positive. Due to this, we include self-connections when computing the score to ensure that the correct time lag is inferred even for the self-connection. For all these test cases, we consider interactions at time lags of zero and one to be the same category as discussed in section \ref{sec3.3}. Here we only tested cMLP, TCDF, GVAR, and JGC as the remaining approaches do not yield information on time lag. The same datasets from the preceding section were used with the exception of fMRI data as it lacks the temporal ground truth, with an additional dataset we will subsequently describe. The resulting scores are collated in Table \ref{table2}.

\subsubsection{VAR and Lorenz96}\label{sec4.2.1}
We see that cMLP, GVAR, and JGC performed well in identifying the correct time lag for these datasets, significantly outdoing TCDF.

\subsubsection{Nonlinear Map}
In addition to the above systems, we simulated a two-dimensional nonlinear map \cite{CCM_w_delay}:
\begin{equation*}
\begin{split}
    x(t) &= x(t-1)[3.78 - 3.78x(t-1) - 0.07y(t-1)] \\
    y(t) &= y(t-1)[3.77 - 3.77y(t-1) - 0.08x(t-\tau)]
\end{split}
\end{equation*}
where in our case we set $\tau=10$. This system was not tested in the preceding section since in the absence of information on time lag, both variables are Granger causal variables of themselves and each other, in which case both AUROC and AUPRC could not be computed.

Table \ref{table2} indicates that cMLP and JGC continue to perform well for this dataset, while GVAR performed comparatively worse next to these two in its AUPRC score. This is likely due to the fact that the nonlinear map contains a nonlinear interaction for two variables at different time lags: one of the terms for $y(t)$ is $y(t-1) \cdot x(t-\tau)$. The formulation of GVAR is such that the model is linear for variables at different time lags, which may then fail to capture this interaction.

\begin{table}[t] 
\caption{Comparison of our proposed approach (JGC) with other existing approaches in identifying Granger causal variables \textit{at the correct interaction time lag} over multiple datasets: VAR, Lorenz96 ($F=10,40$), and nonlinear map.} 
\label{table2} 
\vskip 0.15in 
\centering
\begin{small} 
\begin{sc} 
\begin{tabular}{llcc} 
\toprule 
Dataset & Model & AUROC ($\pm$ SD) & AUPRC ($\pm$ SD) \\ 
\midrule 
\multirow{4}*{VAR}& cMLP & \textbf{1.000 $\pm$ 0.000} & \textbf{0.997 $\pm$ 0.002} \\ 
& TCDF & 0.790 $\pm$ 0.025 & 0.572 $\pm$ 0.056 \\ 
& GVAR & 0.999 $\pm$ 0.001 & 0.985 $\pm$ 0.010 \\ 
& JGC & 0.999 $\pm$ 0.000 & 0.990 $\pm$ 0.005 \\ 
\midrule 
\multirow{4}*{L96$_{F=10}$}& cMLP & 0.953 $\pm$ 0.020 & 0.900 $\pm$ 0.043 \\ 
& TCDF & 0.750 $\pm$ 0.016 & 0.488 $\pm$ 0.034 \\ 
& GVAR & 0.999 $\pm$ 0.000 & 0.981 $\pm$ 0.007 \\ 
& JGC & \textbf{1.000 $\pm$ 0.000} & \textbf{0.999 $\pm$ 0.000} \\ 
\midrule 
\multirow{4}*{L96$_{F=40}$}& cMLP & \textbf{0.999 $\pm$ 0.002} & \textbf{0.993 $\pm$ 0.009} \\ 
& TCDF & 0.696 $\pm$ 0.015 & 0.299 $\pm$ 0.018 \\ 
& GVAR & 0.996 $\pm$ 0.001 & 0.952 $\pm$ 0.012 \\ 
& JGC & 0.992 $\pm$ 0.008 & 0.951 $\pm$ 0.024 \\ 
\midrule 
\multirow{4}*{Map}& cMLP & 0.997 $\pm$ 0.005 & 0.979 $\pm$ 0.043 \\ 
& TCDF & 0.750 $\pm$ 0.000 & 0.533 $\pm$ 0.000 \\ 
& GVAR & 0.952 $\pm$ 0.014 & 0.765 $\pm$ 0.041 \\ 
& JGC & \textbf{1.000 $\pm$ 0.000} & \textbf{1.000 $\pm$ 0.000} \\ 
\bottomrule 
\end{tabular} 
\end{sc} 
\end{small} 
\vskip -0.1in 
\end{table} 

\subsection{Inference of Interaction Sign}\label{sec4.3}
Here we consider the detection of interaction sign in the same manner as \citet{gvar}, through the multi-species Lotka-Volterra system with the following evolution \cite{lotka}:
\begin{align*}
    \frac{d\mathbf{x}^i}{dt} &= \alpha\mathbf{x}^i - \beta\mathbf{x}^i \sum_{j\in Pa\left(\mathbf{x}^i\right)}\mathbf{y}^j-\eta\left(\mathbf{x}^i\right)^2,\ \ \textrm{for } 1\leq i\leq \frac N2,  \\
    \frac{d\mathbf{y}^j}{dt} &= \delta\mathbf{y}^j \sum_{k\in Pa\left(\mathbf{y}^j\right)}\mathbf{x}^k - \rho\mathbf{y}^j,\ \ \textrm{for }1\leq j\leq \frac N2
\end{align*}
where $\mathbf{x},\mathbf{y}$ describe the population of prey and predator species respectively, with $Pa(\cdot)$ containing the indices of the variables in the respective set of Granger causes. The remaining parameter, which modulate the strength of the interactions, were simulated at $\alpha=\rho=1.1,\beta=\delta=0.2,\eta=2.75\times 10^{-5},|Pa\left(\mathbf{x}^i\right)|=|Pa\left(\mathbf{y}^j\right)|=2$ up to $T=2000$ observations, following the parameters given in \citet{gvar}. The system has a dimension $N=20$ consisting of 10 preys and 10 predators. As is expected intuitively, a larger prey population would boost the predator population (positive interaction) while a larger predator population would reduce the prey population (negative interaction). We therefore test if the different appproaches can reveal such a relationship.

To score the approaches in a similar way as the preceding sections, we compute for each interaction sign the sensitivity of each approach, i.e. the proportion of the correct interaction sign identified by each approach. Similar as before, we run each approach over a range of sparsity hyperparameter values and select the best average performance over the average sensitivity and AUPRC. The results are given in Table \ref{table3}, which indicates that most approaches, with the exception of TCDF and eSRU, are able to assign the correct interaction sign for both the positive and negative interactions. We also computed the AUROC and AUPRC for inferring the Granger causal variables, where we see that GVAR and JGC significantly outperform the other models.

\begin{table}[t] 
\caption{Comparison of our proposed approach (JGC) with other existing approaches for the multi-species Lotka-Volterra system. The sensitivity scores for the interaction signs are split into positive and negative interactions.} 
\label{table3} 
\vskip 0.15in 
\centering
\begin{small} 
\begin{sc} 
\begin{tabular}{lcccc} 
\toprule 
Model & AUROC ($\pm$ SD) & AUPRC ($\pm$ SD)\\ 
\midrule 
cMLP & 0.897 $\pm$ 0.044 & 0.404 $\pm$ 0.096 \\ 
cLSTM & 0.777 $\pm$ 0.020 & 0.481 $\pm$ 0.077 \\ 
TCDF & 0.698 $\pm$ 0.032 & 0.172 $\pm$ 0.020 \\ 
eSRU & 0.534 $\pm$ 0.030 & 0.135 $\pm$ 0.022 \\ 
GVAR & \textbf{0.999 $\pm$ 0.001} & 0.993 $\pm$ 0.006 \\ 
JGC & \textbf{0.999 $\pm$ 0.001} & \textbf{0.995 $\pm$ 0.005} \\  
\midrule 
Model & Sensitivity (+) & Sensitivity (-) \\ 
\midrule 
cMLP & 0.980 $\pm$ 0.024 & 0.950 $\pm$ 0.055 \\ 
cLSTM & 0.910 $\pm$ 0.058 & 0.970 $\pm$ 0.040 \\ 
TCDF & 0.970 $\pm$ 0.024 & 0.520 $\pm$ 0.140 \\ 
eSRU & 0.560 $\pm$ 0.107 & 0.650 $\pm$ 0.095 \\ 
GVAR & \textbf{1.000 $\pm$ 0.000} & \textbf{1.000 $\pm$ 0.000} \\ 
JGC & \textbf{1.000 $\pm$ 0.000} & \textbf{1.000 $\pm$ 0.000} \\ 
\bottomrule 
\end{tabular} 
\end{sc} 
\end{small} 
\vskip -0.1in 
\end{table} 

\subsection{Summary of Experimental Results}
The performance across the various test cases is consistently dominated by cMLP, GVAR, and JGC. We show on Table \ref{table4} their AUROC and AUPRC scores averaged over all test cases for inferring Granger causal variables (from Tables \ref{table1} and \ref{table3}) and time lag (from Table \ref{table2}). The comparison over all models are given in the appendix. Here we see that JGC performs best overall.



\begin{table}[t] 
\caption{Average performance of the top three models for Granger causal variable inference and time lag inference over all test cases.} 
\label{table4} 
\vskip 0.15in 
\centering
\begin{small} 
\begin{sc} 
\begin{tabular}{lcccc} 
\toprule 
Mean Score & \multicolumn{2}{c}{Variable} & \multicolumn{2}{c}{Time Lag} \\
Model & AUROC & AUPRC & AUROC & AUPRC\\
\midrule 
cMLP & 0.8924 & 0.7034 & 0.9872 & 0.9670\\
GVAR & 0.9284 & 0.8357 & 0.9864 & 0.9207\\
JGC & \textbf{0.9705} & \textbf{0.8716} & \textbf{0.9989} & \textbf{0.9925}\\
\bottomrule 
\end{tabular} 
\end{sc} 
\end{small} 
\vskip -0.1in 
\end{table} 

\subsection{Stability of the Thresholding Procedures}\label{sec4.5}
In the preceding sections, we restricted the comparison to using AUROC and AUPRC which used only the variable importance scores, and not the inferred Granger causal variables resulting from thresholding the importance scores. For approaches such as cMLP whose results are thresholded by default (due to the training algorithm), the magnitude of the nonzero weights nevertheless served as importance scores.

In this section, we examine the performance and stability of the thresholded results in the top performing methods (cMLP, GVAR, JGC) over the tested range of regularization parameter $\lambda$ for two test systems: VAR and Lorenz96 ($F=40$). Since GVAR has two regularization parameters ($\lambda,\gamma$), we maximize the F-scores over $\gamma$ for each $\lambda$ and inspect how these F-scores change with $\lambda$.

\begin{figure*}[t]
    \centering
    \includegraphics{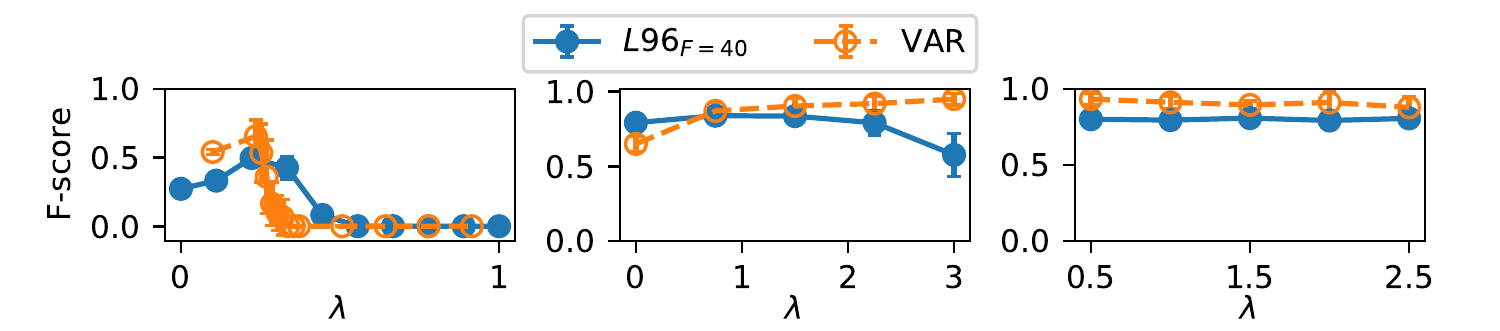}
    \caption{The F-scores for the identification of Granger causal variables across the tested range of hyperparameters for the Lorenz96 ($F=40$) and VAR systems (section \ref{sec4.1}). \textit{Left:} cMLP, \textit{center:} GVAR, \textit{right:} JGC.
    }
    \label{fig:stability}
\end{figure*}

Fig. \ref{fig:stability} shows that both GVAR (center) and JGC (right) are relatively stable across $\lambda$, while cMLP is comparatively sensitive to $\lambda$ which may present difficulty when used in practical settings. In addition, while cMLP achieved high AUROC and AUPRC for both the VAR and Lorenz96 ($F=40$) systems, we see that the F-scores are consistently lower than GVAR and JGC, indicating a weakness in the thresholding procedure for cMLP. While the F-score for GVAR appears to drop dramatically when $\lambda=0$ for one of the test cases, this is not unexpected as the network is completely unconstrained in its representation when $\lambda=0$; what is more important is the stability when $\lambda>0$. For both GVAR and JGC, then, the main consideration is to not set an overly large $\lambda$ which may naturally cause the network to drop the weaker causal variables, but otherwise we see that there is a safe margin for $\lambda$ for these two test cases. The corresponding plots for all the other test cases are given in the appendix.

\section{Experiments on Systems with Nonstationary Structures}\label{nonstationary}
Here we test the performance of our proposed approach for analyzing data whose dependency structures are nonstationary, a phenomenon also known as structural breaks \cite{structural_breaks}. For such systems, the analysis is done in JGC by visualizing the Jacobian, which measures the importance of each variable, as a function of time.

\subsection{Piecewise Vector Autoregression}
\begin{figure}[t]
    \centering
    \includegraphics{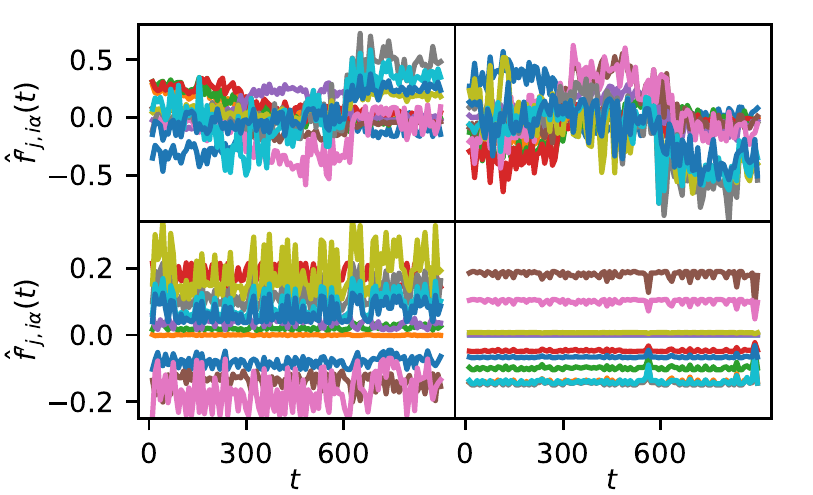}
    \caption{The variable importance measure for two target variables (columns) in the piecewise VAR system with structural breaks at $t=300$ and $t=600$. The bottom row shows how the same quantity would appear if the network is not given time as input.}
    \label{fig:piecewise}
\end{figure}

We first consider VARs defined similarly as \eqref{VAR} in section \ref{varsection}, now with the connectivity matrix $A_\alpha$ being defined piecewise in time. Here we simulated the system with $N=10$ variables and maximum lag $\tau=5$ up to $T=900$, with two structural break points at $t=300,600$ such that $A_\alpha$ abruptly changes in these two time points, being replaced by a matrix with completely different elements. This system therefore has 3 piecewise regimes separated by the break points at $t=300,600$. The results were plotted for two target variables in Figure \ref{fig:piecewise}, showing that our proposed approach is able to detect the 3 piecewise regimes. In the same figure (second row), we show that without providing the network with information on time, this piecewise structure would not be discovered.

\subsection{Historical Stock Prices}
\begin{figure}[t]
    \centering
    \includegraphics{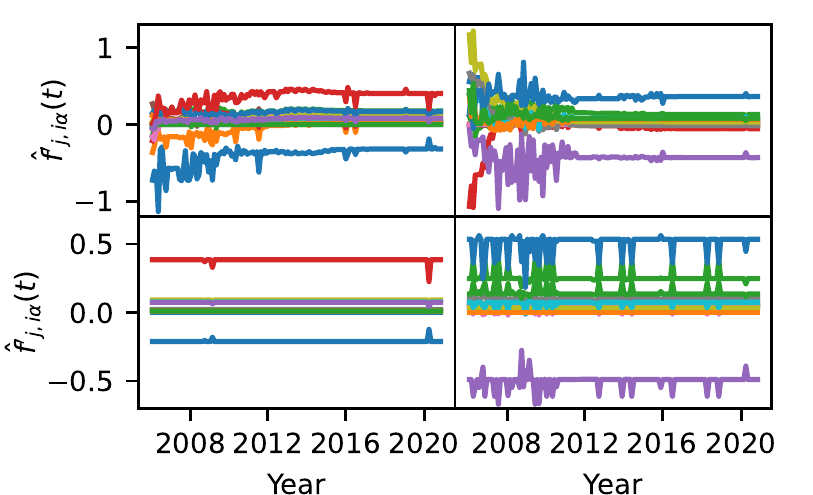}
    \caption{The variable importance measure for two target variables (columns) in the stock returns data. The bottom row shows how the same quantity would appear if the network is not given time as input. For both target variables in the top row, we see at least one variable whose importance measure moves from a nontrivial value towards zero over time.}
    \label{fig:bank}
\end{figure}

We now apply the approach to the weekly log returns for stock prices of banks and financial institutions from Jan 1st, 2006 to Nov 1st, 2020 \cite{bankdata}. This period includes, among others, major events such as the 2007-2008 financial crisis and COVID. Excluding those with incomplete observations, the resulting system has a dimension of $44$ with $751$ observations. Our analysis revealed a mixture of stationary and nonstationary structures for the different target variables in the system. Figure \ref{fig:bank} displays the results for two target variables with nonstationary structures, with particularly striking nonstationarity occuring around the 2007-2008 financial crisis. As in the preceding section, we see that without providing temporal information, the network is unable to learn the nonstationary structures.

\section{Conclusion} 
In this work we introduced JGC, a neural network-based approach for nonlinear Granger causality using the Jacobian as a holistic measure of variable importance. We augmented the Granger causality model with both contemporaneous variables and a measure of time (for nonstationary data). When tested for the identification of Granger causal variables, interaction time lags, and interaction signs, our proposed approach is on par with other approaches across many test systems and outperforms them in some others. In addition, the thresholding procedure for inferring Granger causality was shown to be relatively stable with respect to the regularization parameter. Lastly, we empirically showed that this approach allows one to visualize the instantaneous information flow between variables as a function of time and is capable of detecting time dependencies in the structure in nonstationary systems where the Granger causal structure itself changes with time.

\bibliography{main}

\newpage
\section{Appendix}
\subsection{Stability of Thresholding Procedures}
Here we examine how the stability of the thresholding procedures implemented in cMLP, GVAR, and JGC fare with respect to the regularization parameter $\lambda$ for each of the test cases. In each case, we consider where applicable the inference just the Granger causal variables (section \ref{sec4.1}) and the inference of GC variables at the correct time lag (section \ref{sec4.2}). Interestingly, while cMLP achieves high AUROC and AUPRC scores, we see that the F-scores tend to be significantly worse than GVAR and JGC. We note that the implementation of GVAR does not include a thresholding procedure for inferring the time lag, therefore this aspect could not be evaluated.

\begin{figure}[!h]
    \centering
    \includegraphics{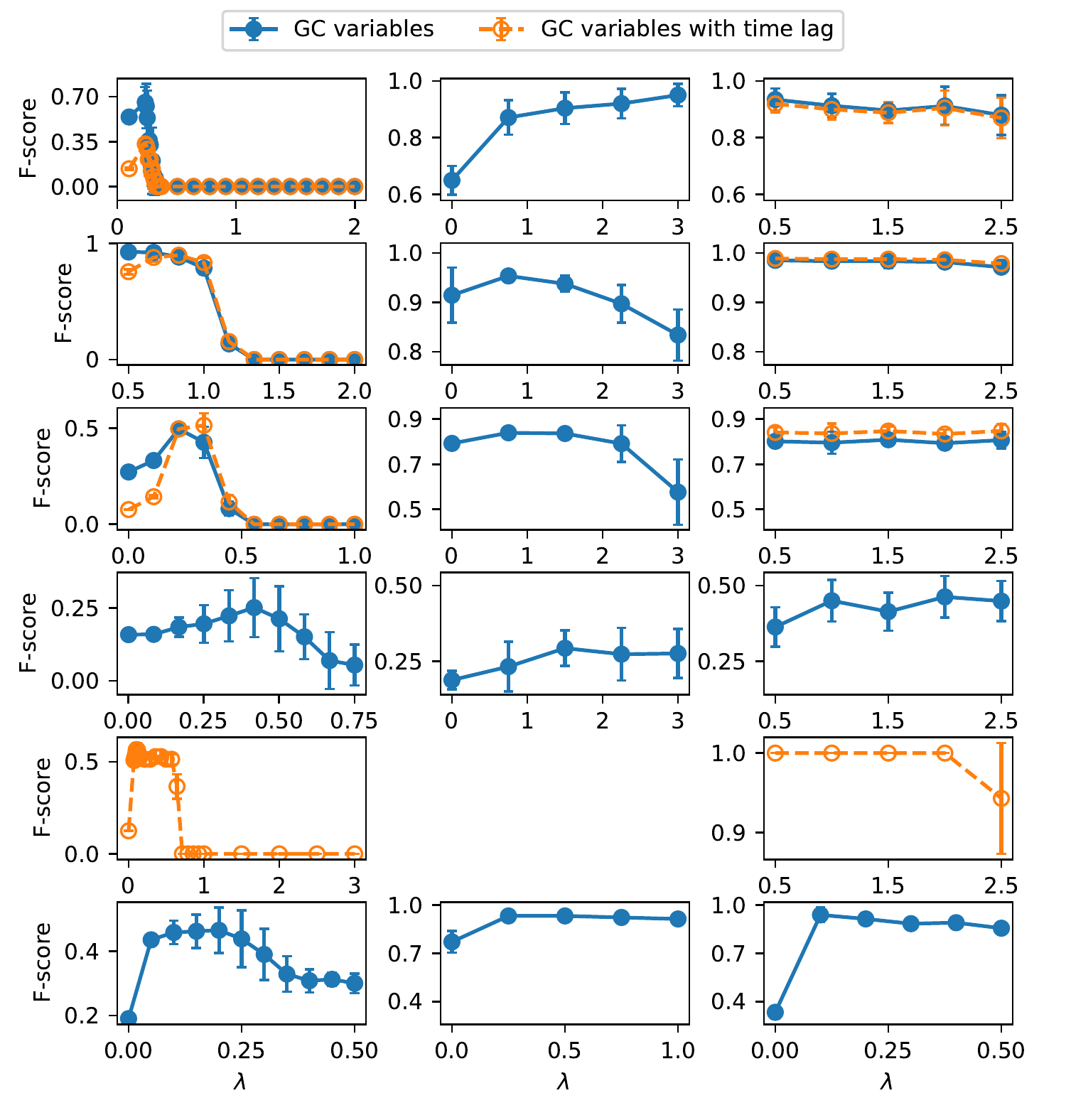}
    \caption{The F-scores achieved by the thresholding procedures of the three top approaches. Columns from left to right correspond to the approach: cMLP, GVAR, and JGC. Rows from top to bottom correspond to the test case: VAR, Lorenz96 ($F=10$), Lorenz96 ($F=40$), fMRI, Nonlinear Map, and LotkaVolterra.}
    \label{fig:appendix}
\end{figure}

\subsection{Summary of experiments for all models}
\begin{table}[H] 
\caption{Average performance of all models for Granger causal variable inference and time lag inference over all test cases.} 
\label{table:average_all} 
\vskip 0.15in 
\begin{center} 
\begin{small} 
\begin{sc} 
\begin{tabular}{lcccccc} 
\toprule 
Variable & cMLP& cLSTM & TCDF & eSRU & GVAR & JGC \\ 
\midrule 
AUROC & 0.8924 & 0.7447 & 0.7901 & 0.7365 & 0.9284 & \textbf{0.9705} \\ 
AUPRC & 0.7034 & 0.5324 & 0.4823 & 0.5426 & 0.8357 & \textbf{0.8716} \\
\midrule 
Lag & cMLP& cLSTM & TCDF & eSRU & GVAR & JGC \\ 
\midrule
AUROC & 0.9872 & - & 0.7465 & - & 0.9864 & \textbf{0.9989} \\ 
AUPRC & 0.9670 & - & 0.4730 & - & 0.9207 & \textbf{0.9925} \\
\bottomrule 
\end{tabular} 
\end{sc} 
\end{small} 
\end{center} 
\vskip -0.1in 
\end{table} 

\subsection{Hyperparameters}
\begin{table}[!h]
\caption{Hyperparameters for the VAR test case.}
\label{var-table}
\vskip 0.15in
\begin{center}
\begin{small}
\begin{sc}
\begin{tabular}{|l||c|c|c|c|c|c|c|}
\hline
Model & \vtop{\hbox{\strut max}\hbox{\strut lag $\eta$}} & \vtop{\hbox{\strut \#hidden}\hbox{\strut \ layers}} & \vtop{\hbox{\strut \#hidden}\hbox{\strut \ units}} & \vtop{\hbox{\strut \#Training}\hbox{\strut \ \ \  epochs}} & \vtop{\hbox{\strut Learning}\hbox{\strut \ \ \ rate}} & Batch size & Sparsity hyperparams \\
\hline
cMLP & 10 & 1 & 100 & 10000 & 1e-2 & NA & $\lambda\in [0.1,2]$ \\
\hline
cLSTM & NA & 1 & 100 & 10000 & 5e-3 & NA & $\lambda\in [0.1,1]$ \\
\hline
TCDF & 10 & 0 & NA & 1000 & 1e-2 & NA & $\alpha\in [0, 2.5]$\\
\hline
eSRU & NA & 2 & 10 & 2000 & 5e-3 & 64 & $\lambda_{1:3}\in [0.01, 0.5]$ \\ 
\hline
GVAR & 10 & 2 & 50 & 1000 & 1e-4 & 64 & $\begin{array}{c}
    \lambda\in [0, 3] \\ 
    \gamma\in [0, 0.025] \\
\end{array}$ \\
\hline
JGC & 10 & 2 & 50 & 2000 & 1e-3 & 64 & $\lambda\in [0.5, 2.5]$ \\ 
\hline
\end{tabular}
\end{sc}
\end{small}
\end{center}
\vskip -0.1in
\end{table}

\begin{table}[!h]
\caption{Hyperparameters for the Lorenz96 ($F=10,40$) datasets.}
\label{lorenz-table}
\vskip 0.15in
\begin{center}
\begin{small}
\begin{sc}
\begin{tabular}{|l||c|c|c|c|c|c|c|}
\hline
Model & \vtop{\hbox{\strut max}\hbox{\strut lag $\eta$}} & \vtop{\hbox{\strut \#hidden}\hbox{\strut \ layers}} & \vtop{\hbox{\strut \#hidden}\hbox{\strut \ units}} & \vtop{\hbox{\strut \#Training}\hbox{\strut \ \ \  epochs}} & \vtop{\hbox{\strut Learning}\hbox{\strut \ \ \ rate}} & Batch size & Sparsity hyperparams \\
\hline
cMLP & 5 & 1 & 100 & 10000 & 1e-2 & NA & $\begin{array}{c}
    \lambda_{F=10}\in [0.5,2] \\ 
    \lambda_{F=40}\in [0,1]
\end{array}$ \\
\hline
cLSTM & NA & 1 & 100 & 10000 & 5e-3 & NA & $\begin{array}{c}
    \lambda_{F=10}\in [0.1,0.6] \\ 
    \lambda_{F=40}\in [0.1,0.25]
\end{array}$ \\
\hline
TCDF & 5 & 0 & NA & 1000 & 1e-2 & NA & $\begin{array}{c}
    \alpha_{F=10}\in [0, 2.5] \\ 
    \alpha_{F=40}\in [0, 2.5]
\end{array}$ \\
\hline
eSRU & NA & 2 & 10 & 2000 & 5e-3 & 64 & $\begin{array}{c}
    \lambda_{1:3,F=10}\in [0.01, 0.1] \\ 
    \lambda_{1:3,F=40}\in [0.01, 0.1]
\end{array}$ \\
\hline
GVAR & 5 & 2 & 50 & 1000 & 1e-4 & 64 & $\begin{array}{c}
    \lambda_{F=10}\in [0, 3] \\ 
    \gamma_{F=10}\in [0, 0.025] \\
    \lambda_{F=40}\in [0, 3] \\
    \gamma_{F=40}\in [0, 0.025]
\end{array}$ \\
\hline
JGC & 5 & 2 & 50 & 2000 & 1e-3 & 64 & $\begin{array}{c}
    \lambda_{F=10}\in [0.5, 2.5] \\ 
    \lambda_{F=40}\in [0.5, 2.5] 
\end{array}$ \\
\hline
\end{tabular}
\end{sc}
\end{small}
\end{center}
\vskip -0.1in
\end{table}

\begin{table}[!h]
\caption{Hyperparameters for the fMRI test case.}
\label{fmri-table}
\vskip 0.15in
\begin{center}
\begin{small}
\begin{sc}
\begin{tabular}{|l||c|c|c|c|c|c|c|}
\hline
Model & \vtop{\hbox{\strut max}\hbox{\strut lag $\eta$}} & \vtop{\hbox{\strut \#hidden}\hbox{\strut \ layers}} & \vtop{\hbox{\strut \#hidden}\hbox{\strut \ units}} & \vtop{\hbox{\strut \#Training}\hbox{\strut \ \ \  epochs}} & \vtop{\hbox{\strut Learning}\hbox{\strut \ \ \ rate}} & Batch size & Sparsity hyperparams \\
\hline
cMLP & 1 & 1 & 50 & 10000 & 1e-2 & NA & $\lambda\in [0.001,0.75]$ \\
\hline
cLSTM & NA & 1 & 50 & 10000 & 1e-2 & NA & $\lambda\in [0.05,0.3]$ \\
\hline
TCDF & 1 & 0 & NA & 1000 & 1e-2 & NA & $\alpha\in [0, 2]$\\
\hline
eSRU & NA & 2 & 10 & 2000 & 1e-3 & 64 & 
$\begin{array}{c}
    \lambda_{1:2}\in [0.01, 0.05]  \\ 
    \lambda_3\in [0.01, 1.0]
\end{array}$ \\
\hline
GVAR & 1 & 2 & 50 & 1000 & 1e-4 & 64 & $\begin{array}{c}
    \lambda\in [0, 3] \\ 
    \gamma\in [0, 0.1] 
\end{array}$ \\
\hline
JGC & 1 & 2 & 50 & 2000 & 1e-3 & 64 & $\lambda\in [0.5, 2.5]$ \\ 
\hline
\end{tabular}
\end{sc}
\end{small}
\end{center}
\vskip -0.1in
\end{table}

\begin{table}[!h]
\caption{Hyperparameters for the nonlinear map test case.}
\label{map-table}
\vskip 0.15in
\begin{center}
\begin{small}
\begin{sc}
\begin{tabular}{|l||c|c|c|c|c|c|c|}
\hline
Model & \vtop{\hbox{\strut max}\hbox{\strut lag $\eta$}} & \vtop{\hbox{\strut \#hidden}\hbox{\strut \ layers}} & \vtop{\hbox{\strut \#hidden}\hbox{\strut \ units}} & \vtop{\hbox{\strut \#Training}\hbox{\strut \ \ \  epochs}} & \vtop{\hbox{\strut Learning}\hbox{\strut \ \ \ rate}} & Batch size & Sparsity hyperparams \\
\hline
cMLP & 15 & 1 & 100 & 10000 & 1e-2 & NA & $\lambda\in [0,3]$ \\
\hline
TCDF & 15 & 0 & NA & 1000 & 1e-2 & NA & $\alpha\in [0, 2.5]$\\
\hline
GVAR & 15 & 2 & 50 & 1000 & 1e-4 & 64 & $\begin{array}{c}
    \lambda\in [0, 3] \\ 
    \gamma\in [0, 0.025] \\
\end{array}$ \\
\hline
JGC & 15 & 2 & 50 & 2000 & 1e-3 & 64 & $\lambda\in [0.5, 2.5]$ \\ 
\hline
\end{tabular}
\end{sc}
\end{small}
\end{center}
\vskip -0.1in
\end{table}

\begin{table}[!h]
\caption{Hyperparameters for the Lotka-Volterra test case.}
\label{map-table}
\vskip 0.15in
\begin{center}
\begin{small}
\begin{sc}
\begin{tabular}{|l||c|c|c|c|c|c|c|}
\hline
Model & \vtop{\hbox{\strut max}\hbox{\strut lag $\eta$}} & \vtop{\hbox{\strut \#hidden}\hbox{\strut \ layers}} & \vtop{\hbox{\strut \#hidden}\hbox{\strut \ units}} & \vtop{\hbox{\strut \#Training}\hbox{\strut \ \ \  epochs}} & \vtop{\hbox{\strut Learning}\hbox{\strut \ \ \ rate}} & Batch size & Sparsity hyperparams \\
\hline
cMLP & 1 & 1 & 100 & 10000 & 5e-3 & NA & $\lambda\in [0, 0.5]$ \\
\hline cLSTM & NA & 1 & 100 & 10000 & 5e-3 & NA & $\lambda \in [0,1]$ \\
\hline TCDF & 1 & 0 & NA & 2000 & 1e-2 & NA & $\alpha\in [0,2]$ \\
\hline eSRU & NA & 2 & 10 & 2000 & 1e-3 & 64 & $\begin{array}{c}
    \lambda_{1:2}\in [0, 0.05]  \\ 
    \lambda_3\in [0, 1.0]
\end{array}$ \\
\hline GVAR & 1 & 2 & 50 & 3000 & 1e-3 & 64 & $\begin{array}{c}
    \lambda\in [0, 1] \\ 
    \gamma\in [0, 0.01] \\
\end{array}$ \\
\hline
JGC & 1 & 2 & 50 & 2000 & 1e-3 & 64 & $\lambda\in [0,0.5]$ \\ 
\hline
\end{tabular}
\end{sc}
\end{small}
\end{center}
\vskip -0.1in
\end{table}

\begin{table}[!h]
\caption{Hyperparameters used in JGC for the nonstationary datasets.}
\label{map-table}
\vskip 0.15in
\begin{center}
\begin{small}
\begin{sc}
\begin{tabular}{|l||c|c|c|c|c|c|c|}
\hline
Dataset & \vtop{\hbox{\strut max}\hbox{\strut lag $\eta$}} & \vtop{\hbox{\strut \#hidden}\hbox{\strut \ layers}} & \vtop{\hbox{\strut \#hidden}\hbox{\strut \ units}} & \vtop{\hbox{\strut \#Training}\hbox{\strut \ \ \  epochs}} & \vtop{\hbox{\strut Learning}\hbox{\strut \ \ \ rate}} & Batch size & Sparsity hyperparams \\
\hline
Piecewise VAR & 10 & 2 & 50 & 2000 & 1e-3 & 64 & $\lambda=3$ \\
\hline Stock Returns & 1 & 2 & 50 & 2000 & 1e-3 & 64 & $\lambda=2$ \\
\hline
\end{tabular}
\end{sc}
\end{small}
\end{center}
\vskip -0.1in
\end{table}

\end{document}